\documentclass[runningheads]{llncs}
\usepackage[T1]{fontenc}
\usepackage{todonotes}
\usepackage{comment}
\usepackage{array}
\usepackage{pifont}
\usepackage{multirow}
\usepackage{graphicx}
\usepackage{hyperref}

\usepackage{color}

\begin{document}
\title{Bench4KE: Benchmarking Automated Competency Question Generation}
\titlerunning{Bench4KE: Benchmarking Automated CQ Generation}
%
\author{Anna Sofia Lippolis\inst{1,2}\orcidID{0000-0002-0266-3452} \and Minh Davide Ragagni \inst{1,3}\orcidID{0009-0009-4332-1780} \and
Paolo Ciancarini\inst{1}\orcidID{0000-0002-7958-9924} \and Andrea Giovanni Nuzzolese\inst{2}\orcidID{0000-0003-2928-9496} \and Valentina Presutti\inst{1,2}\orcidID{0000-0002-9380-5160
}}
\authorrunning{A. S. Lippolis et al.}
%
\institute{University of Bologna, Bologna, Italy \and 
CNR - Institute of Cognitive Sciences and Technologies, Rome \& Bologna, Italy \and University of Pisa, Pisa, Italy 
}
\maketitle              
\begin{abstract}
The availability of Large Language Models (LLMs) presents a unique opportunity to reinvigorate research on Knowledge Engineering (KE) automation. This trend is already evident in recent efforts developing LLM-based methods and tools for the automatic generation of Competency Questions (CQs), natural language questions used by ontology engineers to define the functional requirements of an ontology. However, the evaluation of these tools lacks standardization. This undermines the methodological rigor and hinders the replication and comparison of results. 
To address this gap, we introduce Bench4KE, an extensible API-based benchmarking system for KE automation. 
The presented release focuses on evaluating tools that generate CQs automatically. 
Bench4KE provides a curated gold standard consisting of CQ datasets from 17 real-world ontology engineering projects and uses a suite of similarity metrics to assess the quality of the CQs generated. We present a comparative analysis of 6
recent
CQ generation systems, which are based on LLMs, establishing a baseline for future research. Bench4KE is also designed to accommodate additional KE automation tasks, such as SPARQL query generation, ontology testing and drafting. Code and datasets are publicly available under the Apache 2.0 license.

\centering \textbf{Resource type}: Benchmarking system

\centering \textbf{License}: Apache 2.0

\centering \textbf{DOI}: \href{https://doi.org/10.5281/zenodo.17817277}{10.5281/zenodo.17817277}

\centering \textbf{URL}: \href{https://github.com/fossr-project/ontogenia-cini}{https://github.com/fossr-project/ontogenia-cini}

\end{abstract}
\section{Introduction}
\label{sec:intro}
Competency Questions  (CQs)~\cite{Gruninger1995}, formulated as natural language questions that outline and constrain the scope of an ontology, play a crucial role in Knowledge Engineering (KE). In fact, they are fundamental for defining functional requirements and unit tests~\cite{blomqvist2012ontology} in ontology design processes, such as eXtreme Design~\cite{presutti2009extreme} (XD). In addition, they guide the modeling of an ontology structure by informing the selection of relevant concepts and relationships, and support verification and validation processes of the encoded knowledge~\cite{alharbi2025}.

With the advent of Large Language Models (LLMs), several research efforts aim to reinvigorate research on KE automation, thanks to the models' ability to generate structured representations from natural text. Many tasks have shown the potential of LLM-assisted KE, focusing, for example, on the generation of ontologies from requirements or on the automatic support to requirements elicitation~\cite{lippolis2024ontogenia,lippolis2025assessing}.
Among these, three main approaches: LLM-based generation, reverse-engineering, and retrofitting of CQs, have been identified in the literature~\cite{alharbi2025} and received significant attention~\cite{alharbi2024experiment,antia2023automating,fathallah2024neon,mahlaza-etal-2025-feasibility,pan2024rag,rebboud2024can,zhang2024ontochat}.
These systems rely on LLMs to generate CQs from requirements such as scenarios, or resources such as documents, or from existing ontologies that lack documentation. 
As these tools evolve, the need for systematic and standardized evaluation methods becomes increasingly pressing.  
Although gold standards have been proposed to evaluate automated KE tasks involving CQs, the landscape of KE automation evaluation is scattered and disaligned: to the best of our knowledge, there is still lack of a common reference benchmarking system to support the evaluation of CQ generation systems nor there has been any attempt to rigorously define this task to support advancement of the state of the art through systematic comparative analysis. 
To address this gap, we introduce \textbf{Bench4KE}, a benchmarking system with the overarching aim to systematically evaluate KE automation tasks. BenchKE provides the community with a framework to evaluate the ability of their systems to act as an expert knowledge engineer, who can adapt their skills to any knowledge domain. In its first release, presented in this paper, Bench4KE supports the evaluation of tools that generate CQs. Our benchmark includes gold standard datasets derived from the related literature (see Section \ref{sec:related}) and integrates multiple semantic and lexical similarity metrics to evaluate the semantic proximity and appropriateness of the generated CQs. 

Bench4KE relies on both user stories and heterogeneous source datasets, ontologies, or PDF documents as input. We test six existing tools through Bench4KE, establishing a baseline for future research on the CQ generation task; however, the objective of this work is primarily to launch a platform for future challenge and collaborative research within the Semantic Web community. 

Our key contributions are:
\begin{enumerate}
    \item \textbf{Bench4KE}: An extensible framework for benchmarking KE automation tasks, currently focused on the evaluation of CQ generation systems, along with usage and configuration instructions;
    \item \textbf{Gold Standard Dataset}: A curated and extensible dataset of 843 manually crafted CQs from multiple real-world ontology projects;
    \item A first \textbf{baseline} for the CQ generation task, based on the evaluation of six recent peer-reviewed systems.
\end{enumerate}


The rest of the paper is organized as follows. Section~\ref{sec:related} presents the related work; Section~\ref{sec:thebenchmark} describes Bench4KE, including its usage scenarios, benchmarking dataset, system architecture, evaluation metrics, and execution framework.  Section~\ref{sec:evaluation} shows the evaluation setup and the experimental results. Section \ref{sec:availability} describes the availability of the resource. Finally, Section~\ref{sec:discussion} discusses the evaluation metrics, the impact and extensibility of the benchmark, while Section~\ref{sec:conclusion} concludes the paper.

\section{Related Work}
\label{sec:related}
In this section, we provide an overview of the related works and we explain the novelty of our contributions with respect to: (i) systems for automated CQ generation, (ii) benchmarks for CQ generation with LLMs, (iii) existing approaches to CQ validation; (iv) ontology engineering tools.


\paragraph{\bf Automated CQ Generation.} 
\label{sec:soa-automated-cq-gen}

Automated and semi-automated CQ generation are long-standing tasks \cite{gangemi2022automatically,malheiros2017unification}. Recently, significant effort has been devoted to developing tools that automate CQ generation using LLMs. These systems differ in the type of input they expect, the prompting techniques adopted, and their intended use within the ontology engineering pipeline. 
They can fall into three main approach categories \cite{alharbi2025}: (i) CQ generation; (ii) Reverse engineering of CQs; and (iii) Retrofitting CQs.
Systems that follow the CQ generation approach are the most prevalent. Among them, Ontochat~\cite{zhang2024ontochat,zhao2024improvingontologyrequirementsengineering}, among other KE tasks, supports CQ generation through conversational prompting and is designed to facilitate participatory ontology engineering. NeOn-GPT~\cite{fathallah2024neon} generates CQs from requirements by combining structured prompt templates with ontology metadata. AgOCQs~\cite{antia2023automating,mahlaza-etal-2025-feasibility} is an automatic corpus-based CQ generation tool that uses fine-tuned LLMs to generate CQs from domain-specific corpora. It applies lexical, syntactic, and semantic filters to ensure that generated CQs are well-formed, answerable, and reflect the intended semantics of the ontology. Pan \textit{et al.}~\cite{pan2024rag} introduce a retrieval-augmented generation (RAG) approach to CQ generation. This approach first retrieves relevant ontology content or background knowledge, which is then passed to an LLM to generate context-aware CQs. Rebboud \textit{et al.}~\cite{rebboud2024can} investigate the use of LLMs to generate CQs given an existing ontology, comparing six different models and multiple prompt settings to evaluate how well LLMs can support knowledge engineers in this task. 
RevOnt~\cite{ciroku2024revont} is a CQ reverse engineering system trained on specific domains that extracts CQs directly from knowledge graphs. It operates in three stages: (a) verbalisation abstraction (generalising instance-level triples to class-level text); (b) question generation using a T5-based model; and (c) question filtering via SBERT-based similarity and paraphrase detection, to derive a compact set of core CQs.
For what concerns retrofitting, Alharbi \textit{et al.}~\cite{alharbi2024experiment} introduce the RETROFIT-CQs approach to generate candidate CQs from ontology triples using GPT-3.5 and GPT-4. Di Nuzzo \textit{et al.}~\cite{di2024automated} propose a pipeline that combines LLMs and Knowledge Graphs. 

\paragraph{\bf Benchmarks for CQ generation with LLMs.}
\label{sec:soa-benchmarks} 
Recently, benchmarks have been proposed to evaluate LLM-based tools in KE and cover tasks such as ontology conceptualisation or CQ generation. 

Rebboud \textit{et al.}~\cite{rebboud2024benchmarking} propose a gold standard designed to evaluate how LLMs contribute to ontology conceptualisation tasks, including ontology generation, documentation, and CQ generation. Such a benchmark focuses on understanding the broader modeling capabilities of LLMs. CQsBEN~\cite{alharbi2025} is a gold standard that supports the comparison of manual, semi-automatic, and automatic CQ construction approaches. The dataset is organized around tasks such as requirement scoping and CQ verification, providing guidance for method classification. It is oriented towards mapping and categorizing CQ engineering approaches in the ontology life-cycle.

\paragraph{\bf Approaches to CQ Validation.} 
\label{sec:soa-approaches}

Early tools for CQ validation rely on manually crafted SPARQL queries to check CQs against an ontology.  
Notable examples are \textit{OntologyTest}~\cite{garcia2009ontologytest} or the \textit{Ontology Testing} framework of Blomqvist \textit{et al.}~\cite{blomqvist2012ontology}.  
Other solutions, such as \textit{CQChecker}~\cite{bezerra2013cqchecker} and the \textit{CQ-Driven Authoring and Testing} plugin by Dennis \textit{et al.}~\cite{dennis2017computing}, integrate CQ validation directly into the ontology‐development workflow.  
Another solution is \textit{TESTaLOD}~\cite{Carrriero2019} that enables agile knowledge graph testing by providing an automated, adaptable framework based on a Web interface a REST API for assessing the quality and consistency of knowledge graphs via CQ validation, error provocation, and logic consistency checking. Instead, \textit{CLaRO}~\cite{keet2019claro} introduces a controlled natural language for authoring CQs, to assist ontology engineers in formulating requirements. 
Finally, Wiśniewski \textit{et al.}~\cite{wisniewski2019analysis} distilled a set of recurring CQ templates.
Traditional systems are evaluated on benchmark ontologies by measuring (i) the accuracy of answering CQs or detecting violations and (ii) coverage. Some studies complement these metrics with user experiments to verify that automated checks align with human judgments of ontology completeness (e.g., Dennis \textit{et al.}~\cite{dennis2017computing}).

LLMs remove the manual formalisation step by mapping a natural-language CQ directly to a query or answer.  
Tufek \textit{et al.}~\cite{tufek2024validating} compare LLM-generated CQs with expert gold standards to assess precision, recall, and F\textsubscript{1}.  
Alharbi \textit{et al.}~\cite{alharbi2024experiment} rely on semantic similarity; Pan \textit{et al.}~\cite{pan2024rag} emphasize precision and answer consistency; and Di Nuzzo \textit{et al.}~\cite{di2024automated} use cosine similarity between expected and produced answers along with the LLM-based rating of each CQ on a scale from 1-5 for what concerns relevance, clarity, and depth.  
Although Lippolis \textit{et al.}~\cite{lippolis2025assessing} focus on ontology generation, they likewise treat CQs as a benchmark of completeness. Compared with earlier work, recent evaluations go beyond simple answer accuracy: researchers now analyze retrieval metrics for generated queries, semantic‐similarity scores, and expert assessments in addition to the traditional coverage measure.
\vspace{-1em}

\paragraph{\bf Ontology Engineering Tools.}
\label{sec:soa-onteng-tools}

Bench4KE is broadly introduced as an ontology engineering tool in which the evaluation of CQ generation systems is only one step of the ontology engineering benchmarking pipeline. Recently, the need for automation with LLM has garnered interest in creating such tools.
The already cited OntoChat~\cite{zhang2024ontochat} is a conversational ontology engineering framework that uses an LLM-based agent to elicit and manage requirements in various stages until ontology generation. 
NeOn-GPT~\cite{fathallah2024neon} is a pipeline that integrates the NeOn ontology development methodology with GPT-based prompts to convert domain descriptions into OWL/Turtle ontologies. 

In summary, existing studies have paved the way for reinvigorating research on knowledge engineering automation, but they are hardly comparable. 
However, a reusable end-to-end benchmark that automatically supports the evaluation of the semantic quality of generated CQs is missing and is arguably needed to systematically advance the state-of-the-art in this area. Bench4KE fills this gap by pairing a gold standard set of 843 CQs manually generated by expert ontology engineers from 17 real‑world ontologies with a validation pipeline that combines classical similarity metrics and an LLM‑based semantic judge, delivering the first systematic, extensible framework for comparing LLM‑driven CQ generation tools.

\section{Bench4KE}
\label{sec:thebenchmark}
 We present Bench4KE by detailing: (i) the requirements specification; (ii) the usage scenarios considered for evaluating the task of CQ generation; (iii) the benchmark dataset used as gold standard; (iv) the system architecture; (v) the execution setup for CQ validation, and (vi) the evaluation metrics supported.

\begin{figure}[h]
    \centering
    \includegraphics[width=1.0\textwidth]{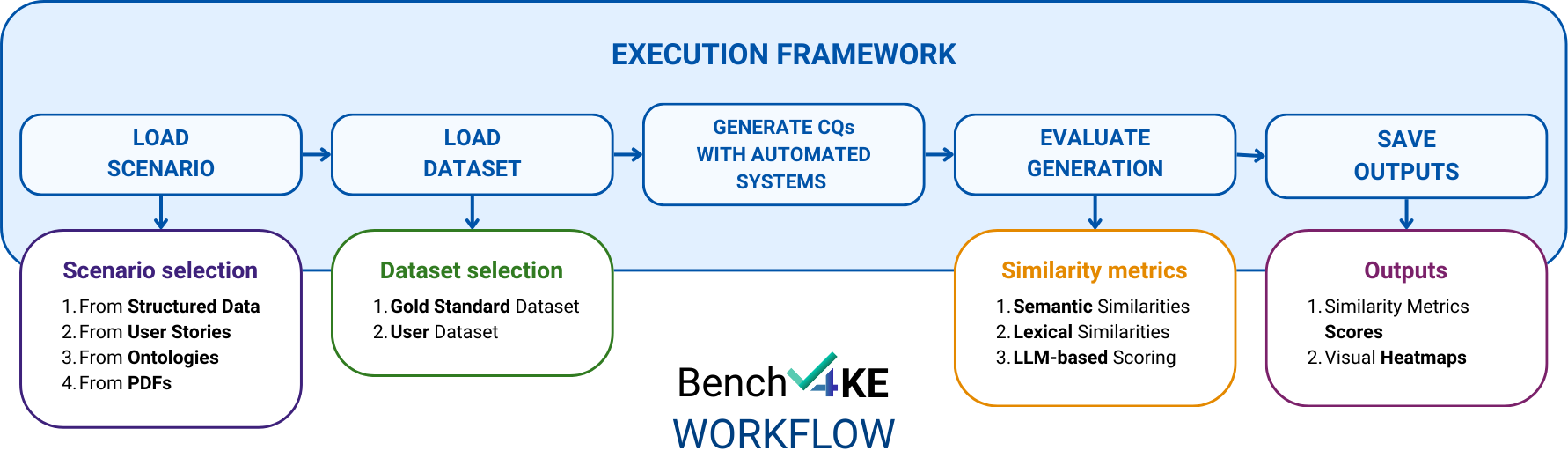}
    \caption{Bench4KE's workflow for CQ validation.}
    \label{fig:benchmark_workflow}
\end{figure}

\begin{table}[ht]
\centering
\caption{Functional (F) and non functional (N) requirements for Bench4KE. The Proven Feature column represents the successfully built and tested functionalities; the Current Deficit column shows areas that need future development.}
\label{tab:bench4ke_requirements}
\scriptsize
\renewcommand{\arraystretch}{1.25}
\setlength{\tabcolsep}{4pt}
\begin{tabular}{|p{0.5cm}|p{7cm}|>{\centering\arraybackslash}p{2cm}|>{\centering\arraybackslash}p{2cm}|}
\hline
\textbf{} & \textbf{Requirement} & \textbf{Proven feature} & \textbf{Current deficit} \\
\hline
F1 & The system shall support the evaluation of CQ generation tools based on user story inputs. & \ding{51} & \\
F2 & The system shall support the evaluation of CQ generation tools using raw structured data. & \ding{51} & \\
F3 & The system shall support the evaluation of CQ extraction or retrofitting tools from ontologies. & \ding{51} &  \\
F4 & The system shall support the evaluation of CQ generation tools from academic PDF sources. & \ding{51} &  \\
F5 & The system shall allow external datasets to be uploaded as gold standards for benchmarking. & \ding{51} & \\
F6 & The system shall expose a RESTful API for dataset access and benchmark submission. & \ding{51} & \\
F7 & The system shall support multiple LLM models for validation. & \ding{51} & \\
F8 & The system should provide configurable prompt templates for domain-specific CQ generation contexts. &  & \ding{51} \\
F9 & The system should support incremental/streaming CQ generation with partial result visibility. & \ding{51} &  \\
N1 & The system shall ensure standardized formats and traceability of results. & \ding{51} & \\
N2 & The system shall support the comparative benchmarking of multiple CQ generation systems. & \ding{51} & \\
N3 & The system shall provide transparent evaluation metrics and benchmarking reports. & \ding{51} & \\
N4 & The system will provide ontology coverage analytics comparing CQ sets against ontology terms. &  & \ding{51} \\
\hline
\end{tabular}
\end{table}

\vspace{-3em}

\subsection{Requirements specification}

Requirements are a prerequisite for successful software development~\cite{Wiegers2013}. For Bench4KE, we derived requirements from related work (§ \ref{sec:related}) to build on validated ideas, expose missing capabilities as requirements, and provide a focused direction for future development.

The elicited requirements capture the core features, functionalities, and quality characteristics expected from Bench4KE as a benchmarking system. They serve as a guide for us and for other developers who want to extend or reuse Bench4KE, supporting the systematic development of a robust and maintainable benchmark. We document both functional and non-functional requirements using the established MASTER template and The SOPHISTs \cite{rupp2020requirements}. This template distinguishes the degree of obligation: ``shall'' marks mandatory requirements, ``should'' marks desirable but optional improvements, and ``will'' marks planned extensions beyond the current scope.

Table \ref{tab:bench4ke_requirements} summarizes the requirements for Bench4KE. Functional requirements (F) describe the essential capabilities of the Bench4KE system, while non-functional requirements (N) capture performance, usability, and other quality criteria. The Proven Feature column indicates which requirements are already implemented or empirically validated, and the Current Deficit column highlights open points and planned enhancements for future releases. Together, these dimensions support progress monitoring and gap analysis, ensuring that Bench4KE evolves in line with user needs and benchmarking goals. In line with this specification, Bench4KE already fulfills most of the defined functional and non-functional requirements, which demonstrates its suitability for a broad range of evaluation scenarios.

\subsection{Usage scenarios}
\label{sec:bench4ke-usage-scenarios}
Bench4KE is designed to support the evaluation of KE automation tasks. In this paper, we present a release focused on LLM-based CQ generation. Although our approach is not specific to any particular type of tool, we currently focus on LLM-based tools, as they represent the most recent and prominent approach in the field. We consider two main scenarios that are faced in real-world ontology projects: i) the existence, as input, of structured or semi-structured data that shall be conceptually and syntactically homogenized (e.g. datasets, documents, ontologies, etc.), which is common in data integration tasks, and ii) the elicitation of requirements through user stories, a common situation in most ontology development methodologies. These two scenarios are usually combined. User stories are crucial to understanding the domain and identifying the boundaries of ontological requirements. Existing data are to be analyzed (via reverse engineering) in order to uncover their underlying conceptual model, which is often implicit. These usage scenarios are designed to test the generalizability of CQ generation models across different input modalities.
\subsection{Benchmarking dataset} 
\label{sec:bench4ke-dataset}
We built a gold standard dataset by collecting CQs from 17 real-world ontology-driven projects, combining various knowledge domains, as reported in Table~\ref{tab:project-table} along with statistics. Each CQ is accompanied by a resource related to one or more usage scenarios. As a result, we obtain 843 CQs. 

\begin{table}
\centering
\caption{Bench4KE dataset statistics.}
\label{tab:project-table}

\resizebox{0.9\textwidth}{!}{%
\begin{tabular}{|l|l|c|c|c|c|c|}
\hline
\textbf{Project} & \textbf{Domain(s)} & \textbf{CQs} & \textbf{Stories} & \textbf{Datasets} & \textbf{Ontologies} & \textbf{PDFs} \\
\hline
\textit{Polifonia}~\cite{de2023polifonia} & Music & 67 & \ding{51} &  &  &  \\
 & Music & 28 &  &  & \ding{51} &  \\
\hline
\textit{ArCo}~\cite{carriero2019arco} & Cultural Heritage & 10 &  & \ding{51} &  &   \\
\hline
\textit{WHOW}~\cite{lippolis2025water} & Water & 9 &  & \ding{51} &  &  \\
 & Health & 6 &  & \ding{51} &  &  \\
\hline
\textit{HACID}~\cite{Kurvers2023,trianni2023hybrid} & Health & 7 & \ding{51} &  &  &  \\
 & Climate Services & 2 &  & \ding{51}  &  & \\
\hline
\textit{SWO}~\cite{malone2014} & Life Sciences & 88 &  &  & \ding{51} &  \\
\hline
\textit{Stuff}~\cite{keet2014} & Macroscopic Stuff & 11 &  &  & \ding{51} &  \\
\hline
\textit{AWO}~\cite{keet2020} & African Wildlife & 14 &  &  & \ding{51} &  \\
\hline
\textit{Dem@Care}~\cite{fernandez2019} & Health & 107 &  &  & \ding{51} &  \\
\hline
\textit{OntoDT}~\cite{panov2015} & Computer Science & 14 &  &  & \ding{51} &  \\
\hline
\textit{Wine Ontology}~\cite{noy2001} & Wine & 7 &  &  & \ding{51} &  \\
\hline
\textit{DOREMUS}~\cite{achichi2018} & Music & 58 &  &  & \ding{51} &  \\
\hline
\textit{NORIA-O}~\cite{tailhardat2024} & ICT & 26 &  &  & \ding{51} &  \\
\hline
\textit{Odeuropa}~\cite{lisena2022} & Cultural Heritage & 74 &  &  & \ding{51} &  \\
\hline
\textit{VGO}~\cite{fernandez2019,parkkila2017} & Videogames & 68 &  &  & \ding{51} &  \\
\hline
\textit{VICINITY Core}~\cite{fernandez2019} & IoT & 126 &  &  & \ding{51} &  \\
\hline
\textit{HCI}~\cite{karras2023} & Hum.-Comp. Interaction & 15 &  &  &  & \ding{51} \\
\hline
\textit{RE}~\cite{costa2022} & Requirement Engineering & 106 &  &  &  & \ding{51} \\
\hline
\multicolumn{2}{|l|}{\textbf{Total}} & \textbf{843} & 74 & 27 & 621 & 121  \\
\hline
\end{tabular}%
}
\end{table}

\label{subsec:architecture}
\subsection{Bench4KE architecture and execution flow}
\label{sec:bench4ke-architecture}
Bench4KE is designed to systematize the evaluation of automatically generated CQs against a curated gold standard dataset. It follows an API-centric architecture and is implemented using the modular and extensible FastAPI\footnote{\url{https://fastapi.tiangolo.com/}} framework.

\begin{figure}[!ht]
    \centering
    \includegraphics[width=0.85\textwidth]{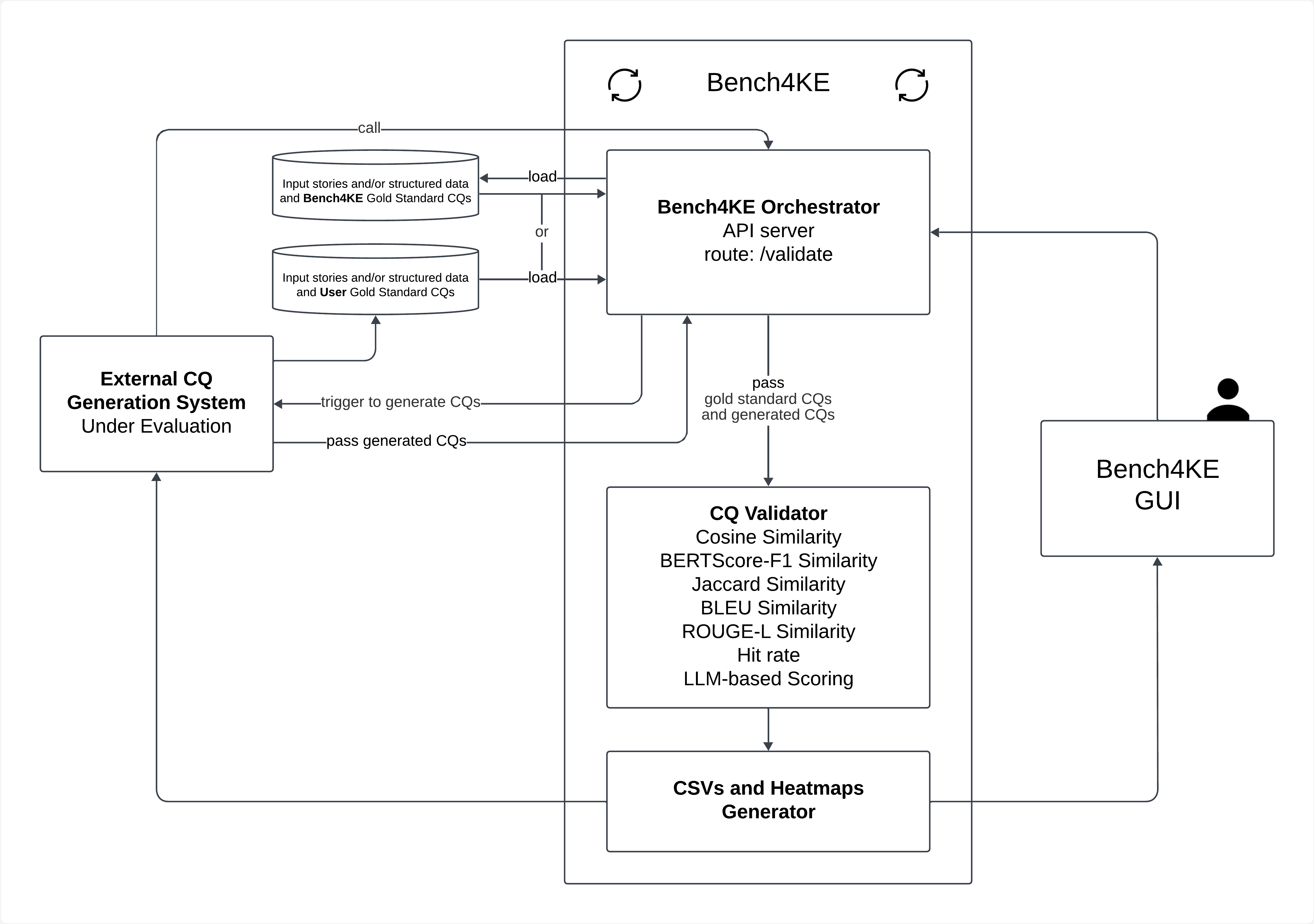}
    \caption{Bench4KE's system architecture.}
    \label{fig:architecture}
\end{figure}

Figure~\ref{fig:architecture} shows the system's architecture, structured to support comprehensive benchmarking capabilities with a focus on maintainability and scalability. 

\paragraph{\bf Validator.} The implementation mirrors the modular layout shown in Figure~\ref{fig:architecture}. The Bench4KE Validation API exposes a single REST endpoint, \texttt{/validate}, implemented through FastAPI. During an evaluation session, the external system (i.e. the system under evaluation) calls the Bench4KE Orchestrator service to initiate the session. Optionally, the caller may provide a custom input dataset, along with a corresponding set of gold standard CQs. This dual capability ensures flexibility, supporting both predefined and ad hoc evaluation scenarios tailored to experimental needs.
Then, the Orchestrator triggers the external system to generate the CQs. Once generated, the CQs are returned to the Orchestrator, which forwards both the generated and gold standard CQs to the CQ-Validator. The Validator computes evaluation scores (§\ref{sec:evaluation}). Currently, the Validator API is not available online, but can be run locally. The following cURL command launches a complete validation call:

{\scriptsize
\begin{verbatim}
curl -X POST "http://127.0.0.1:8000/validate/" \
  -F "use_default_dataset=true" \
  -F "external_service_url=http://127.0.0.1:8001/newapi" \
  -F "api_key=your_key" \
  -F "validation_mode=all" \
  -F "model=gpt-4" \
  -F "save_results=true"
\end{verbatim}
}

The parameters are defined as follows:
{\scriptsize
\begin{itemize}
    \item \verb|use_default_dataset|: Set to \verb|true| to use the benchmark dataset.
    \item \verb|external_service_url|: The URL of the external CQ generation API.
    \item \verb|api_key|: If required by the external service.
    \item \verb|validation_mode|: One of: \verb|all|, \verb|cosine_bertscore_judge|, \verb|llm|, \verb|cosine|, \verb|jaccard|.
    \item \verb|output_folder|: Directory where the outputs and heatmaps are stored.
    \item \verb|model|: LLM to use.
    \item \verb|save_results|: If set to \verb|true|, stores the results as CSV.
    \item \verb|save_every|: It is set to a number for incremental savings.
\end{itemize}
}

\label{subsec:evaluation-metrics}

\paragraph{\bf Evaluation metrics.} The platform enables the execution of multiple evaluation strategies that assess both syntactic and semantic correspondence between automatically generated and manually authored CQs. The evaluation results are presented both numerically and visually. The following evaluation metrics are currently implemented: Cosine similarity (computed over sBERT~\cite{Reimers2019} sentence embeddings), BERTScore~\cite{Zhang2019BERTScoreET}, Jaccard similarity, ROUGE-L~\cite{Lin2004}, BLEU~\cite{Papineni2002}, and LLM-based semantic scoring. We also implement a metric called \textit{Hit Rate}, where the validator counts how many gold CQs obtain at least one generated counterpart whose Cosine similarity exceeds the 0.60 threshold. Hit rate is therefore a coverage indicator: high values mean that the generator reliably produces at least one relevant CQ per requirement.

Hit Rate, along with other metrics, collectively helps quantify both surface-level and structural overlaps between generated and gold-standard CQs. In fact, BERTScore captures contextual and distributional semantics, Jaccard focuses on token-level overlap, while BLEU and ROUGE-L assess n-gram precision and recall. Nevertheless, these metrics often fail to reflect deep semantic equivalence, especially when paraphrasing is involved. 
To mitigate this issue, we enhance the analysis with an LLM-based semantic scoring that interprets the metric outputs and selects the most semantically aligned CQ pairs.
The prompt asks the model to identify the closest matching CQs and to highlight any essential questions missing from the gold standard\footnote{\url{https://github.com/fossr-project/ontogenia-cini/blob/main/restapi/app/services/cq_validator.py}.}. Apart from that, we add an LLM-as-a-judge metric derived from \cite{di2024automated}, where the LLM is asked to judge the responses on a scale from 1 to 5 on the parameters of relevance, clarity and depth. This approach allows Bench4KE to benefit from the reproducibility of lexical metrics, while introducing semantic depth and human-like reasoning through LLM analysis. For what concerns LLMs, we support multiple models, including those supported by OpenAI\footnote{\url{https://openai.com/}}, Anthropic\footnote{\url{https://www.anthropic.com/}}, Meta\footnote{\url{https://ai.meta.com/meta-ai/}}, and Together AI\footnote{\url{https://www.together.ai/}}. 



\paragraph{\bf Output Structure.}
The evaluation outputs are automatically stored in a user-defined output directory as JSON and visualized on the console.

\subsubsection{Interface.}
To facilitate human interaction, Bench4KE includes two additional software components: (i) a CQ generation app\footnote{\url{https://github.com/fossr-project/ontogenia-cini/blob/main/restapi/cq_generator_app.py}}, which simulates the behavior of a CQ generation system and its communication with Bench4KE, and (ii) a web front-end\footnote{\url{https://github.com/fossr-project/ontogenia-cini/blob/main/restapi/bench4ke-validate-ui.py}} that allows users to run the evaluation entirely on their local machine, returning the results in JSON format. The front-end accepts as input an external service URL. Once provided, the system computes the metrics and displays them immediately in the browser; the same results (together with heatmap images) are written to a dedicated output directory for later inspection.


\subsection{Documentation}
Comprehensive and detailed documentation can be accessed from the Github repository. The documentation is organized into the following main parts: \textit{Getting Started} and \textit{How to Use?}, respectively, one with all the necessary information to help users begin using Bench4KE, and includes installation instructions and a quickstart guide and helping both creators of CQ generation systems and general users of the system who might want to test or customize the tool with the information to make Bench4KE work. 



\section{Evaluation and impact}
\label{sec:evaluation}
To test Bench4KE as a benchmarking system for CQ generation, and in line with our goal of evolving it as a community-driven effort, we involved community researchers who have been actively working on automatic CQ generation. We contacted the authors of seven recent peer-reviewed solutions that were not domain-specific and who made the code available~\cite{antia2023automating,di2024automated,fathallah2024neon,pan2024rag,mahlaza-etal-2025-feasibility,rebboud2024can,zhang2024ontochat}, focused on LLM-based CQ generation. Six expressed interest in testing the tool. Then, we evaluated the integration capabilities of these systems with Bench4KE.
To establish a first baseline, apart from these systems, we have selected the \texttt{cosine\_bertscore\_judge} configuration, i.e., cosine similarity over SBERT embeddings, BERTScore-F1, and an LLM-as-a-judge score. This choice is motivated not only by computational considerations but also by the complementarity of these metrics and their suitability for the CQ generation task. Furthermore, through discussion and a Survey (cf. Section \ref{sec:impact}) we have been able to assess Bench4KE's potential impact, improve it, and adjust future implementation plan.

\subsection{Analyzed systems}

The current baseline is given from the existing systems interested in testing our tool and from the set of datasets of the systems themselves used for the single evaluations, along with European project resources which we used in the beta version of Bench4KE and with which we have extended the dataset.

To achieve integration of Bench4KE with existing systems, minor modifications were necessary to some of them. In particular, OntoChat required limiting the output to a maximum number of CQs, as the tool requires this setting, and allowing to inject also PDF, ontology, and dataset text instead of user stories; the generation from user stories was preserved. NeOn-GPT originally was evaluated on the Wine Ontology, as its specific use case. Therefore, the authors proposed adapting the prompt to allow a more general evaluation of the generated CQs. They also slightly modified the output generation as it included additional text along the CQs, which would result in injecting noise in the metrics computation. In the case of the work by Pan \textit{et al.}, we also expanded the system so that it could detect basing on the format type of the input whether to use RAG (on PDFs) or chat (in the other cases).

Similarly, AgOCQs also required some input reorganisation and adjustments to the requested output format. RETROFIT-CQs' few-shot templates were updated to reflect the details of each ontology analyzed in the benchmarking system. Details of the changes for each system may be included in their respective projects' Github. We consider these modifications and the potential limitation they imply as part of the process of standardisation, as these systems have been developed with different scenarios and project goals in mind and evaluated in isolation with custom gold standards. They allow us to provide a reference baseline for the CQ generation task and identify the variations of this task that shall be considered in future developments of Bench4KE.

All evaluations have been executed on the same laptop with Intel(R) Core(TM) i5-8265U CPU @ 1.60GHz 1.80 GHz, with 8Gb of RAM. For the LLM, when no explicit specification was set, we use GPT-4o as it was the most used LLM in the related literature, with temperature set to 0.

\vspace{-1em}
\subsection{Results}
\label{sec:results}


\begin{table}[h!]
    \centering
    \caption{Performance comparison across usage scenarios on Bench4KE.
    Approaches are distinguished according to the classification proposed by~\cite{alharbi2025} based on two classes, i.e. Generating and Retrofitting. 
    Similarity metrics range 0 to 1, with Cosine similarity, computed on sBERT embeddings, and BERTScore-F1, balancing token-level precision and recall by rewarding both correct matches and adequate coverage of the reference answer. 
    LLM-as-a-Judge scores represent averaged quality ratings on a 1--5 scale.}
    \label{tab:performance}

\renewcommand{\arraystretch}{1.25}
    \setlength{\tabcolsep}{6pt}
\resizebox{0.9\textwidth}{!}{%
    \begin{tabular}{
        p{3.1cm}   
        p{1.9cm}   
        | c c      
        | c c c    
    }
        \hline

        & &
        \multicolumn{2}{c|}{\textbf{Similarity}} 
        & \multicolumn{3}{c}{\textbf{LLM-as-a-Judge}} \\
        \cline{3-4} \cline{5-7}

        \textbf{System} & 
        \textbf{Approach} &
        \textbf{Cosine} &
        \textbf{F1} &
        \textbf{Relev.} &
        \textbf{Clar.} &
        \textbf{Depth} \\

        \hline

        \textit{OntoChat}~\cite{zhang2024ontochat}       & Generating & 0.24 & 0.60 & 4.47 & 4.70 & 3.52 \\
        \textit{NeOn-GPT}~\cite{fathallah2024neon}       & Generating & 0.24 & 0.58 & 4.63 & 4.65 & 3.90 \\
        \textit{AgOCQs}~\cite{mahlaza-etal-2025-feasibility} & Generating &   0.16   &    0.60  &   3.60   &     4.10 &  2.50    \\
        \textit{Pan et al.}~\cite{pan2024rag}            & Generating & 0.32 & 0.60 & 4.78 & 4.54 & 3.90 \\
        \textit{Rebboud et al.}~\cite{rebboud2024can}    & Generating & 0.30 & 0.60 & 4.67 & 4.55 & 3.98 \\
        \textit{RETROFIT-CQs}~\cite{li2025retrofit}      & Retrofitting & 0.21 & 0.57 & 4.49 & 4.48 & 3.97 \\

        \hline
    \end{tabular}
    }
\end{table}

\begin{figure}[!htb]
    \centering
    \includegraphics[width=0.65\linewidth]{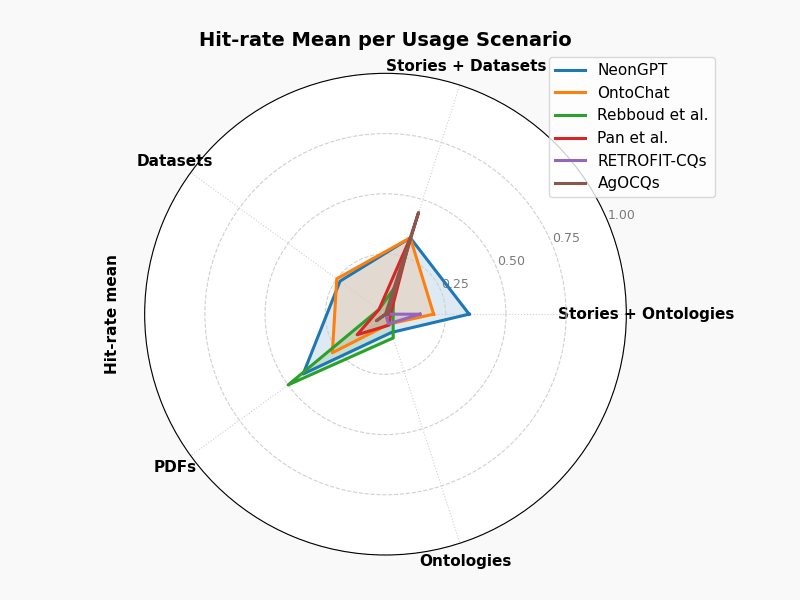}
    \caption{Mean Hit Rate across systems according to the usage scenarios.}
    \label{fig:hitrate}
\end{figure}

Table \ref{tab:performance} reports the performance of the six evaluated systems using the Cosine similarity, BERTScore and LLM-as-a-judge metrics with a total average across projects. In Figure \ref{fig:hitrate}, we also provide an overview of the systems with respect to the contemplated usage scenarios. These results provide a first baseline; however, it has to be considered that we hope for community participation to add new datasets and domains so as to obtain a more robust baseline.
The low absolute values observed may sometimes be due to either the fact that the systems were evaluated on a usage scenario or file formats the system was not tested on, or a possible limitation of similarity-based metrics in evaluating questions that might be fit but are not similar to the gold standard CQs. This topic is addressed in Section \ref{sec:discussion} of the paper. 


\subsection{Impact of Bench4KE}
\label{sec:impact}

To test the beta version of the benchmark, the authors of three of the systems involved in our experiments provided feedback through a survey, including an assessment of the usefulness and usability of Bench4KE and possible suggestions for its future improvement. The survey was adapted from the System Usability Scale~\cite{Brooke1996}. It included a series of statements in which the raters express their agreement on a scale from 1 (strongly disagree) to 5 (strongly agree), covering the system’s usefulness, extensibility, support for user-provided datasets, clarity of instructions, and usability of the interface. The Survey questionnaire's responses are available on \href{https://github.com/fossr-project/ontogenia-cini/blob/main/experimental-results/Bench4KE's%20Evaluation%20(responses).csv}{GitHub} and the questionnaire itself is online for future collection of users' feedback on Bench4KE. All users agreed that Bench4KE fills a critical gap by providing a community reference for evaluating CQ generation tools and is a vital contribution towards advancing research in this domain. Users stated that they plan to use Bench4KE to evaluate methods/tools for automatic CQ generation to evaluate other KE automation tasks that will be supported. They perceive that the system is easy to use, including its graphical user interface. The main suggestion for improvement is on the metrics supported by Bench4KE to assess the performance of the tool. These suggestions have been implemented in the first  Bench4KE release, which we present in this paper.

\vspace{-1em}


\section{Availability, sustainability and licensing}
\label{sec:availability}

Benck4KE is available on a public \href{https://github.com/fossr-project/ontogenia-cini}{GitHub repository}, which aggregates all its material (software code, data, experiment configurations, results, etc.), and it is archived on \href{https://zenodo.org}{Zenodo}, which provides its \href{https://doi.org/10.5281/zenodo.17817277}{DOI: 10.5281/zenodo.17817277}. Bench4KE is released under the \href{https://www.apache.org/licenses/LICENSE-2.0}{Apache 2.0 License}. The documentation accompanying each release reduces the barrier for first‑time users. \\
\textbf{Sustainability} is guaranteed by the University of Bologna and CNR‑ISTC for at least five years; additionally, Bench4KE is intended to solicit community effort and has already received the interest and willingness to contribute from researchers external to the authors' group and institutions. Both \href{https://github.com/fossr-project/ontogenia-cini/blob/main/maintenance.md}{maintenance plan} and \href{https://github.com/fossr-project/ontogenia-cini/blob/main/contributionguidelines.md}{contribution guidelines} are available on Github.\\
\textbf{Standards compliance}. Bench4KE exposes a REST interface that returns JSON‑LD and CSV. Dataset metadata in DCAT-AP is also available online.

\section{Discussion}
\label{sec:discussion}
In this section, we discuss the envisioned current and future implications of the resource for the Semantic Web community.

\subsection{Beyond similarity. How should CQs be evaluated?}

CQ elicitation inherently requires human involvement. We do not consider automatic evaluation to be sufficient or exhaustive; rather, we regard it as a useful but limited support tool. The metrics we report follow conventions established in the related literature and should be understood as approximations rather than definitive judgements of quality. LLM-based scoring can contribute to this evaluation pipeline, especially for the evaluation of novelty in the generated CQs. For example, since in our system the LLM proposes possible additional questions to add, these can, upon scrutiny, be incorporated into an evolving set of gold standard CQs, effectively simulating an enrichment process. With the aim of enriching the gold standard dataset, CQs that do not match the gold standard ones but are sufficiently similar to the LLM-proposed ones can also be considered in the Hit Rate scoring and eventually added as gold standard. Over time, this iterative refinement can help us move towards a more comprehensive coverage of the CQ space, while keeping human judgement at the center of the evaluation.

\subsection{Registered Interest from the Semantic Web Community}

The authors of existing CQ generation systems expressed interest in Bench4KE, including, for some of them, the willingness to participate in its future development. The systems evaluated with Bench4KE are now showing an ``Evaluated with Bench4KE'' text in their project's official README file (e.g. see \url{https://github.com/King-s-Knowledge-Graph-Lab/OntoChat/tree/main}). This is a first step towards a shared methodological approach to define and evaluate knowledge engineering automation tasks. Feedback from a first round of user testing supports the relevance and usability of Bench4KE. The respondents agreed that the system is a valuable reference for evaluating CQ generation tools and a promising foundation for future KE automation tasks. They highlighted the importance of extending the support to additional metrics and expressed a strong interest in using the system for future research.

\subsection{Novelty of Bench4KE}
To date, there is no dedicated benchmarking system for evaluating KE automation at large, making Bench4KE a novel contribution. Although gold standards for CQ validation exist, they are not tailored to the challenges and characteristics of LLM-generated outputs. Also, most of existing gold standards are use cases- and domain-specific. Bench4KE is designed to evaluate a system over multiple domains and use case scenarios. Moreover, coordinated, community-wide evaluation efforts in this space are lacking. Bench4KE aims to address this gap by providing a standardized and extensible evaluation framework, which can incorporate and evolve with datasets and gold standard resources from diverse projects and knowledge domains.  

\subsection{Extensibility}
The framework is extensible in three main directions: (i) support for additional datasets, (ii) extension to additional metrics, and (iii) extension to other Knowledge Engineering tasks. Regarding datasets, users can either evaluate their own datasets locally or request integration into the official Bench4KE dataset following the contribution guidelines outlined in \url{https://github.com/fossr-project/ontogenia-cini/blob/main/contributionguidelines.md}. The latter requires meeting specific quality criteria to ensure consistency. Although the system is already comprehensive for what concerns the implemented metrics, additional ones that also value the LLM-generated questions that do not match with the gold standard ones are a planned improvement, which will be first discussed within a community effort.
As for other KE tasks, future work will involve expanding Bench4KE into a suite of REST APIs that support the broader ontology development life cycle. According to user feedback on the first release (§\ref{sec:impact}), Bench4KE could be extended beyond CQ validation: respondents expressed interest in new KE automation tasks such as CQ-to-SPARQL mapping, ontology alignment, ontology requirements extraction, and entity recognition from domain texts. These suggestions point to the potential for Bench4KE to evolve into a comprehensive evaluation suite for various stages of KE workflows.

\subsection{Limitations and Future Work} 

In this moment, the baseline we propose in this paper has some limitations. What is present in the benchmark default dataset corresponds to what exists in the related works, which by default is imbalanced; however, the system can currently be configured to be more balanced and extensible with additional datasets. The variety of dataset formats also remains limited to XML, JSON, and CSV. Secondly, there is a risk of data leakage. This problem is relevant, and some works in the state of the art are addressing it, e.g. \cite{li2025retrofit,paulheim2025ontologies}, for instance by running models without internet access or using generated benchmarks on the fly. In the future, we plan to add to the system these solutions for data leakage.

Future work also includes extending the dataset with more data formats and usage scenarios, issuing a call for community collaboration to contribute on GitHub, and promoting wider adoption and a more robust baseline through a future challenge and through its integration into Knowledge Engineering courses at the University of Bologna. Additional developments will also focus on expanding coverage across more stages of the ontology engineering process.



\section{Conclusion}
\label{sec:conclusion}
In this paper, we introduced Bench4KE, a benchmarking system that in this release evaluates tools that automatically generate Competency Questions. Bench4KE fills a critical gap, as it is intended to homogenize the evaluation of KE automation tasks and to ease comparative analysis: the existing effort has indeed been evaluated in isolation against project-specific gold standards. In general, Bench4KE is designed to evolve to support the evaluation of additional knowledge engineering automation tasks, such as SPARQL query generation, ontology generation, ontology testing, etc. With 17 gold standards from real-world ontology projects covering several domains, and its ability to load custom gold standards, Bench4KE can be used to evaluate CQ generation systems on their ability to act as expert knowledge engineers, while also recognizing their strengths within specific domains. We establish a baseline for the LLM-based CQ generation task by performing a comparative analysis of six recently published systems. The results show that this task is challenging. Researchers from other institutions than those of the authors have provided useful feedback, including expressing willingness to contribute and use Bench4KE in their future research. They evaluated the system as a necessary resource, finding it easy to use.

\begin{credits}
\subsubsection{\ackname} We gratefully acknowledge the authors of NeOn-GPT, RETROFIT-CQs, OntoChat, AgOCQs, the work by Pan \textit{et al.} and Rebboud \textit{et al.} for their interest in our work, contribution, and collaborative exchanges. 
\end{credits}
\\
{\small
\textbf{Use of Generative AI.} ChatGPT was used to enhance the readability of some of the text and improve the language of this paper, after the content was first added manually. After using this service, the author(s) reviewed and edited the content as needed and take(s) full responsibility for the content of the published article.
}

\bibliographystyle{splncs04}
\bibliography{bibliography}

\end{document}